%% file: main.tex
\documentclass{article}

\PassOptionsToPackage{numbers, compress}{natbib}

\usepackage[final]{neurips_2020}

\usepackage[utf8]{inputenc} 
\usepackage[T1]{fontenc}    
\usepackage{hyperref}       
\usepackage{url}            
\usepackage{booktabs}       
\usepackage{amsfonts}       
\usepackage{nicefrac}       
\usepackage{microtype}      
\usepackage[super]{nth}

\usepackage{makecell}
\usepackage{graphicx}
\graphicspath{ {./figures/} }

\title{Hyper-optimization with Gaussian Process and Differential Evolution Algorithm}

\author{%
Jakub Klus
\texttt{jakub.klus@innovatrics.com} \\
Innovatrics s.r.o \\
\And
Pavel Grunt
\texttt{pavel.grunt@innovatrics.com} \\
Innovatrics s.r.o \\
\And
Martin Dobrovolný
\texttt{martin.dobrovolny@innovatrics.com} \\
Innovatrics s.r.o \\
}

\begin{document}

\maketitle

\begin{abstract}
Optimization of problems with high computational power demands is a challenging task.
A probabilistic approach to such optimization called Bayesian optimization lowers performance demands by solving mathematically simpler model of the problem.
Selected approach, Gaussian Process, models problem using a mixture of Gaussian functions.
This paper presents specific modifications of Gaussian Process optimization components from available scientific libraries.
Presented modifications were submitted to BlackBox 2020 challenge, where it outperformed some conventionally available optimization libraries.
\end{abstract}

\input{introduction}
\input{methods}
\input{results}
\section{Conclusion}
This paper showed several modifications to an available implementation of Gaussian Process optimization procedure.
Proposed modifications were evaluated on a test set provided by BlackBox 2020 challenge organizers.
The evaluation showed that at least one example submission outperformed selected approach.
This was also confirmed by the preliminary BlackBox 2020 challenge leaderboard, where the selected submission landed around \nth{45} place.

However, at the final evaluation (disclosed to BlackBox 2020 challenge participants), the selected submission achieved the \nth{11} place outperforming example submissions greatly.
The best explanation we can give as authors is that kernel discretization can play a major role, especially for the case of categorical input parameters.
This assumption is further supported by the fact that there were no assignments with categorical inputs in the available test set.

\section*{Broader Impact}
Authors believe that this work will motivate contributors and maintainers of black-box optimization scientific packages to incorporate proposed changes into their software. Our optimization framework is available at \url{https://github.com/Brown-Box/brown-box}.

\begin{ack}
This work was not directly funded by any funding organization.
However, authors would like to thank selected open-source projects, which made the submission possible, namely: \textit{scikit-learn}\,\cite{scikit-learn} and GNU \textit{parallel}\,\cite{gnu_parallel}.
\end{ack}

\bibliography{references}

\end{document}

%% file: introduction.tex
\section{Introduction}
The recent popularity of Machine Learning (ML) methods broadens its applicability in many research fields.
Selecting the best ML method and tuning its parameters is a common approach to achieve state-of-the-art results.
But, with rising data size and method complexity, parameter optimization (so-called hyper-optimization) quickly exhausts available computational power. 
For example, the systematic optimization of deep learning (DL) training parameters\,\cite{DBLP:journals/corr/abs-1803-09820, 10.5555/3042817.3042832} can take several days to perform.

The time required for optimization is proportional to the number of iterations we perform (number of samples taken).
The number of samples can be significantly lowered if we model the relation between individual parameters and outputs (so-called objective function).
A probabilistic framework for the hyper-optimization is called Bayesian optimization\,\cite{10.1023/A:1008306431147}.
Bayesian optimization defines a surrogate model of the objective function and loss function, which defines how optimal are future queries.

Shahriari \textit{et al.}\,\cite{7352306} provide a comprehensive review of Bayesian optimization applications and available implementations.
Moreover, three basic methods of Bayesian optimization are described in\,\cite{7352306}, namely: Gaussian Process (GP), Tree of Parzen Estimators (TPE), and Random forests (RF).

A GP (also described in\,\cite{rasmussen2005gaussian}) approximates objective function with a mixture of multivariate Gaussian functions.
The loss of future queries is then calculated from the predicted mean value and confidence.
On the contrary, the objective function in TPE (described thoroughly in \cite{10.5555/2986459.2986743}) is modelled by two hierarchical processes.
One process describes parameters yielding better values, and the second describes parameters yielding worse values.
Loss is then evaluated using the ratio of these two hierarchical processes.
In RF regression a surrogate model consisting of an ensemble of simpler models made on random subsets of problems is constructed.
A GP approach was selected for this paper since it provides more space for experimental improvements as described in the following sections.

\subsection{Kernel function}

To estimate Gaussian mixture coefficients efficiently, covariances between individual Gaussian components are calculated using kernel trick\,\cite{rasmussen2005gaussian}.
Selection of kernel is one of the crucial parts of GP, individual kernel functions and even their arithmetic are described in\,\cite{duvenaud_2014,rasmussen2005gaussian}
Commonly used kernels are squared exponential kernel and Matérn 5/2 kernel \cite{10.5555/2999325.2999464}.
Both aforementioned kernels implement automatic relevance determination (ARD).
Briefly, ARD works by maximizing the marginal likelihood of the GP regression model with respect to the kernel length-scale parameters.
If the length-scale corresponding to the objective function parameter is large, the model is almost independent of it.

\subsection{Non-continuous parameters}
\begin{figure}[!ht]
\centering
\includegraphics[width=0.99\textwidth]{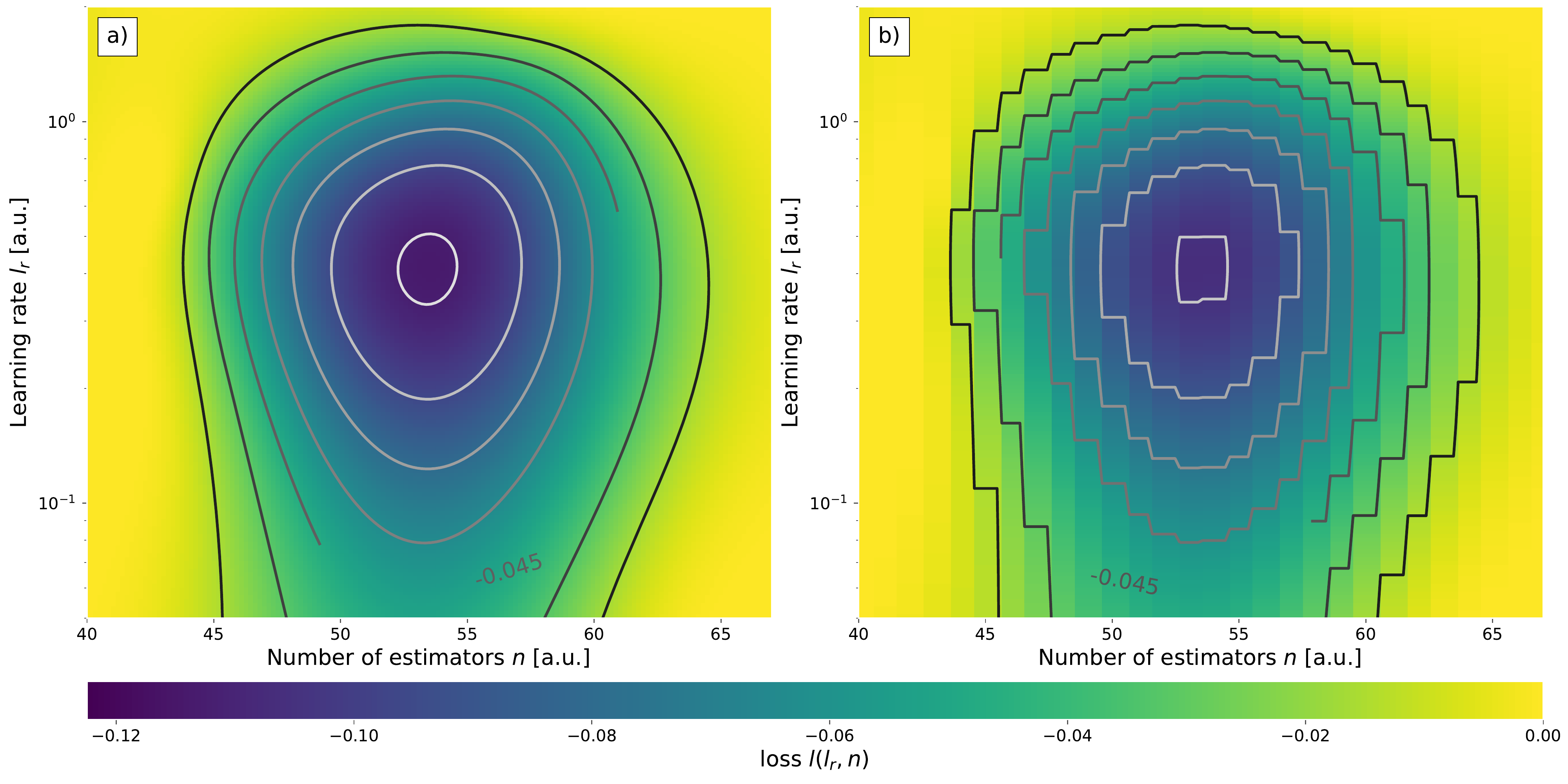}
\caption{
Comparison of loss function for a) kernel function without transformation, b) for kernel function with transformation  (see section \ref{subsec:discrete_input_space} for implementation details).
}
\label{fig:ei_discretization}
\end{figure}

A GP needs to be extended to work with non-continuous parameters like categories, integer numbers, or true/false values.
Therefore, Garrido-Merchán and Hernández-Lobato in\,\cite{GARRIDOMERCHAN202020} proposed an extension of GP for non-continuous domains.
They discourage the naive approach of converting parameters to the real domain and rounding outputs of GP afterwards.
To truly model non-continuous parameters they suggest to modify kernel function and apply rounding inside of it.
This ensures that interactions between individual parameters will not be extrapolated to undefined values (\textit{e.g.} real values between integers).
The difference of these approaches is depicted in Figure \ref{fig:ei_discretization}.
The practical implementation is discussed in section \ref{subsec:discrete_input_space}.

\subsection{Parallelization}
Bayesian optimization belongs to a family of sequential-model based optimization (SMBO) methods.
In SMBO, a new query is calculated whenever a model prediction is updated with data acquired from the previous step.
A GP must be modified to enable parallel evaluation of queries in batches.
In \cite{Ginsbourger2010} two basic methods of parallelization were proposed: \textit{constant liar} and \textit{kriging believer}.
In both methods, a batch of queries is calculated sequentially by predicting the result of the first query and inserting it into the GP model.
For the \textit{constant liar} approach, known points are expanded by constant $L$ corresponding to the parameters of the previous query.
For the \textit{kriging believer} approach, known points are expanded by the current GP model mean corresponding to the parameters of the previous query.
The full potential of parallelization was exploited in\,\cite{wang2019parallel}, where authors sampled multiple queries from the loss function directly.

\subsection{Challenge details}
Algorithms described in this paper were submitted to the Black-Box Optimization for Machine Learning challenge (BlackBox 2020 challenge), a part of NeurIPS 2020 Competition Track.
In short, the optimization was performed in 16 batches of 8 queries, and optimizers executing longer than 640 seconds were stopped.
Moreover, the BlackBox 2020 challenge organizers defined a test set of assignments and kept a preliminary leaderboard of submissions.
Organizers also provided a compilation of example submissions utilizing selected optimization libraries and means of comparison using a defined score.

%% file: methods.tex
\section{Methods}\label{sec:methods}
The proposed solution is a combination of methods described in the previous section built upon a \textit{scikit-learn} library\,\cite{scikit-learn}.
Scikit-learn provides an implementation of GP regression and basic kernels, including kernel arithmetic.
The selected loss function was Expected Improvement (EI)\,\cite{10.1023/A:1008306431147} defined as follows:
\begin{equation}\label{eq:loss_function}
    EI=\left(f_{\textrm{min}} - \hat{y}\right)\Phi\left(\frac{f_{\textrm{min}}-\hat{y}}{s}\right)+s\phi\left(\frac{f_{\textrm{min}}-\hat{y}}{s}\right),
\end{equation}
where $f_{\textrm{min}}$ is the best observed value, $\hat{y}$ is the value predicted by GP model, $s$ is the standard deviation of the model prediction, $\Phi\left(\cdot\right)$ is a standard normal distribution function, and $\phi\left(\cdot\right)$ is a standard normal density function.

The parallelization was implemented according to the \textit{kriging believer} approach since the \textit{constant liar} is straightforward and sampling of multiple queries is complicated.
\subsection{Input space transformations}\label{subsec:discrete_input_space}
The nature of objective function parameters can affect the performance of the GP model.
Therefore it is advisable to transform the aforementioned parameters into domains where their interactions are balanced.
Such domain transforms can incorporate changes proposed in \cite{GARRIDOMERCHAN202020}.
Numerical values are transformed with respect to their "configuration space" (\textit{e.g.} parameters with "log" configuration space are transformed using $\log_{10}$ function).
Categorical values are transformed to the range $[0.0, 1.0]$ using one-hot encoding, thus increasing the input dimension for GP.
Boolean true and false values are transformed to $1.0$ and $0.0$ respectively.

The kernel function transformation is implemented using type coercion.
Real numbers corresponding to integer objective function parameters are coerced by rounding.
Real numbers from non-linear space are rounded to the nearest transformed integer (\textit{e.g.} parameters from "log" configuration space are coerced by $\log_{10}(\lfloor 10^{x}\rceil))$).
Each real number vector corresponding to a categorical value is modified by replacing its maximal value by $1.0$ and the rest of its values by $0.0$.
Real numbers corresponding to Boolean values are rounded and clipped to range $[0.0, 1.0]$.

\subsection{Initialization}
The GP has to be primed with at least two queries. 
With respect to the parallelization, the number of prime queries has to be multiple of batch size.
As only the objective function parameter ranges and spaces are known, the simplest option to initialize the GP is a batch of randomly generated parameter values.
More advanced initialization mitigating possibilities of sampling nearly identical random parameters is Latin hypercube (LH) initialization\,\cite{doi:10.1080/01621459.1998.10473803}.
In the LH initialization, we create a set of possible prime queries using a uniform grid over all parameters.
The uniform used by the LH is constructed in the transformed input space.
Afterwards, we randomly sample unseen points from the set of possible prime queries to create a batch.

\subsection{Meta-optimizer}
New queries from GP correspond to minimums in loss function~(\ref{eq:loss_function}).
We call the optimization inside Bayesian optimization a \textit{meta-optimization} and the selecting optimizer a \textit{meta-optimizer}.
We utilized methods provided in the \textit{sckit-learn} library, namely the default optimizer L-BFGS (an approximation of  Broyden–Fletcher–Goldfarb–Shanno algorithm\,\cite{10.5555/39857}) and implementation of a non-gradient method, Differential Evolution\,\cite{10.1023/A:1008202821328}.
The Differential Evolution algorithm was chosen according to the investigation of transformed $EI$ loss (see $EI$ loss calculated on an example problem selected from the test set in figure \ref{fig:ei_discretization}b) because L-BFGS can have lower performance in non-continuous spaces.
The negative impact of discretization on L-BFGS meta-optimizer is confirmed experimentally, see Table \ref{tab:meta-optmizers}.

%% file: results.tex
\section{Results and Discussion}
Proposed changes were evaluated on a test set given by organizers of BlackBox Challenge 2020.
Initially, a problem of domain discretization with respect to meta-optimizers primed by two random batches was investigated.

Results in Table~\ref{tab:meta-optmizers} show that Differential Evolution meta-optimizer performs better for complex kernel approximation.
In contrast, the score of L-BFGS-B meta-optimizer for complex kernel transformation is significantly lower than the score for naive transformation.
In sum, L-BFGS meta-optimizer outperforms Differential Evolution, but cannot beat example submissions.

\begin{table}[!ht]
\caption{Performance of different meta-optimizers with respect to the discretization regime.
Proposed modifications of domain discretizations are evaluated on two meta-optimizers and compared with example submissions (pySOT, turbo, and hyperOpt).
The score is calculated from BlackBox Challenge 2020 test set.
}
\label{tab:meta-optmizers}
\centering
\begin{tabular}{lllr} 
 \toprule
 Optimizer & Meta-optimizer	& Discretization & Score \\
 \midrule
 pySOT      & \quad                     & \quad     & 98.385 \\ 
 turbo      & \quad                     & \quad     & 97.660 \\ 
 Proposed   & L-BFGS-B                  & naive     & 97.316 \\ 
 Proposed   & L-BFGS-B                  & complex   & 96.556 \\ 
 Proposed   & Differential Evolution    & complex   & 96.296 \\
 hyperOpt   & \quad                     & \quad     & 96.147 \\ 
 Proposed   & Differential Evolution    & naive     & 95.957 \\ 
 \bottomrule
\end{tabular}
\end{table}

We tried to exploit potential benefits of employing Differential Evolution meta-optimizer by improving its initialization.
Apart from the initialization type (discussed in Section~\ref{sec:methods}) a size of initialization batch was concerned.
Initialization batch size increment is motivated by the increase of the probability of hitting a promising parameter configuration.
The last examined modification was initialization of meta-optimizer, denoted as \textit{meta-initialization}.
Random meta-initialization can sustain the same drawbacks as in the case of initialization.
Therefore, we proposed a quasi-random meta-initialization that samples objective function parameters from the vicinity of known promising points.

Results in Table~\ref{tab:init-params} show how can the initialization improve the performance of proposed optimizer.
It can be stated that a larger number of priming batches helps in most cases, where initialization was not done purely randomly.
Also, the non-random initialization improves the resulting score significantly.
The combination of the Differential Evolution meta-optimizer, the complex discretization of the kernel, LH initialization, and quasi-random meta-initialization was chosen as the final submission for the BlackBox 2020 challenge.

\begin{table}[!ht]
\caption{
Score values for different initialization and meta-initialization options.
The evaluated proposal combines the Differential Evolution meta-optimizer and the complex discretization.
The quasi-random meta-initialization method samples the objective function parameters from the vicinity of known promising points. 
}
\label{tab:init-params}
\centering
\begin{tabular}{llclr} 
 \toprule
 Optimizer & Initialization & \makecell{Initialization \\ size [batch]} & Meta-initialization & Score \\
 \midrule
 pySOT      & \quad     & \quad & \quad         & 98.385 \\
 Chosen     & LH        & 5     & quasi-random  & 97.871 \\
 Proposed   & random    & 5     & quasi-random  & 97.464 \\ 
 Proposed   & LH        & 2     & quasi-random  & 97.309 \\ 
 Proposed   & random    & 2     & quasi-random  & 97.028 \\ 
 Proposed   & LH        & 2     & random        & 95.957 \\ 
 Proposed   & random    & 5     & random        & 96.671 \\ 
 Proposed   & LH        & 5     & random        & 96.450 \\ 
 Proposed   & random    & 2     & random        & 96.296 \\ 
 \bottomrule
\end{tabular}
\end{table}